\documentclass[letterpaper, 10 pt, conference]{ieeeconf}  

\IEEEoverridecommandlockouts                              

\usepackage{multirow}
\usepackage{svg}

\usepackage{graphicx} 
\usepackage{amsmath,amssymb,amsfonts}
\usepackage{hyperref}

\usepackage{cite}

\usepackage{subcaption}

\usepackage{booktabs}
\usepackage{color}

\title{\LARGE \bf
DRL-Based Pose Control for Double-Ackermann
Robots Under Actuation Uncertainties
}

\author{ Oussama Zaim$^{1*}$, Mélodie Daniel$^{1*}$, Aly Magassouba$^{2}$, Miguel Aranda$^{3}$, and Olivier Ly$^{1}$
\thanks{*These authors contributed equally.}
\thanks{$^{1}$Univ. Bordeaux, CNRS, Bordeaux INP, LaBRI, UMR 5800, F-33400 Talence, France. $^{2}$School of Computer Science, University of Nottingham,
UK. $^{3}$Instituto de Investigación en Ingeniería de Aragón (I3A), Universidad de Zaragoza, 50018 Zaragoza, Spain. Author’s Accepted Manuscript. Released under the Creative Commons license: Attribution 4.0 International (CC BY 4.0).  Corresponding author: Mélodie Daniel, e-mail: \texttt{melodie.daniel@u-bordeaux.fr.}}}%

\begin{document}

\maketitle
\thispagestyle{empty}
\pagestyle{empty}


 \begin{abstract}    
Robust deployment of deep reinforcement learning (DRL) policies on real robots remains challenging due to discrepancies between simulation and real-world dynamics. We address this issue in the context of maneuvering with double-Ackermann-steering mobile robots, which introduce additional constraints due to their non-holonomic nature. Building upon the DRL framework ManeuverNet, we extend its objective from position control to full pose control, resulting in a more challenging task. We further investigate the impact of actuation-related uncertainties on policy transfer. The use of simplified actuation models during training of the extended policy can lead to poor generalization, shown by a success rate drop from 100\% in PyBullet to 25\% in Gazebo under stricter evaluation conditions. To address this limitation, we adopt a sim-to-sim-to-real approach, where actuation effects observed in Gazebo are incorporated into the PyBullet training environment. Using multi-environment DRL with SAC and CrossQ, we learn policies that remain robust despite modeling inaccuracies. This approach can significantly reduce the performance gap across simulators, achieving up to 92\% success rate in Gazebo and maintaining 69\% under stricter thresholds, with successful transfer to a real robot without additional tuning. 

\end{abstract}

 \section{Introduction}
The successful use of deep reinforcement learning (DRL) has enabled significant progress in robotic control, particularly for navigation and locomotion tasks involving omni-directional~\cite{Mehmood2021ITC}, quadruped~\cite{Zhang2024AS}, or biped robots~\cite{GaspardIROS2024}. In the case of Ackermann-steering vehicles~\cite{Zhao2013AC}, their strong kinematic constraints require coordinated forward and rotational motion and raise specific challenges~\cite{Siegwart2005}. In this work, we focus on double-Ackermann-steering mobile robots (DASMRs), a class of four-wheel-steering (4WS) platforms that improve maneuverability and stability compared to single-Ackermann systems~\cite{Yu2010TRO}. However, this added flexibility increases control complexity, as both front and rear steering must be coordinated. As a result, executing precise maneuvers, such as alignment or parking, remains challenging, especially in confined environments~\cite{ICRA2026}.

Classical planning and control methods have been widely applied to DASMRs. Approaches such as Timed Elastic Band (TEB)~\cite{TEBref} can generate feasible trajectories under kinematic constraints. However, they rely on accurate system modeling and a substantial tuning effort, often for each specific robot and operating condition~\cite{TEB_parameters_limitation}. Consequently, their performance is highly sensitive to modeling errors, noise, and variations in actuation behavior, limiting their robustness in real-world deployments~\cite{ICRA2026, Arce2023}.

\begin{figure}[t]
    \centering
    \captionsetup{font=scriptsize}
    \includegraphics[width=\linewidth]{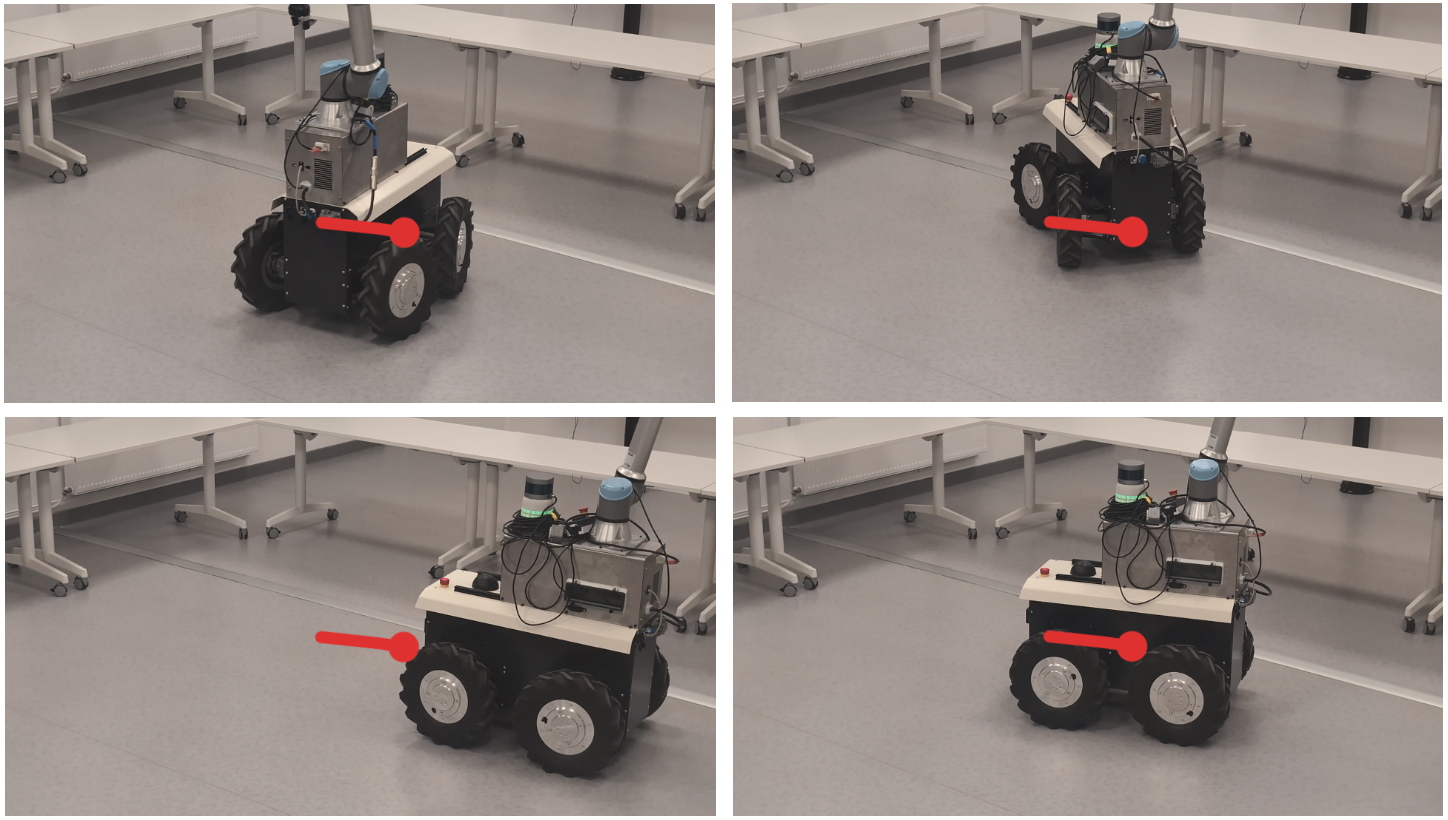}
    \put(-240,76){A}
    \put(-240,7){C}
    \put(-117,76){B}
    \put(-117,7){D}
    \vspace{2pt}
    \caption{Trajectory example illustrating maneuver execution with a 4WS robot. The images from A to D show snapshots of the trajectory from start to finish. The red sphere represents the desired position, while the red bar indicates the desired orientation.}\vspace{-0.6cm}
    \label{fig:placeholder}
\end{figure}
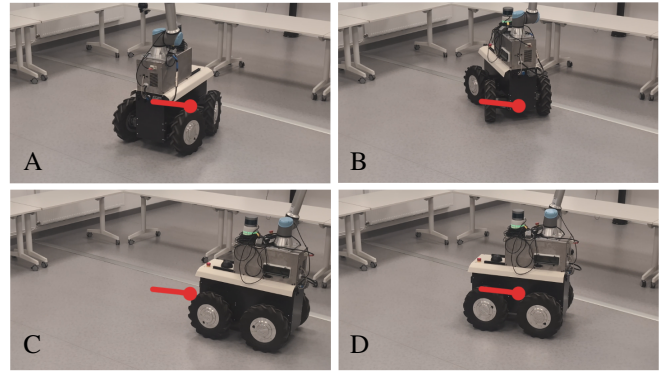

Deep reinforcement learning (DRL) offers an alternative by learning control policies directly from interaction, reducing the need for explicit modeling and manual tuning. This approach has been explored in works such as ManeuverNet~\cite{ICRA2026}, which focuses on maneuvering in obstacle-free environments, as well as in DRL-based local navigation methods that handle obstacles~\cite{Daniel2025ManeuverNetWithObstacles}. While the position control enabled by these prior approaches is highly efficient and successful in relevant tasks, in certain practical scenarios (e.g., parking, docking to charging stations, or alignment for manipulation in agricultural contexts) it is also useful to explicitly control the vehicle's final heading; i.e., to perform maneuvers with full pose control.




Bridging the gap between simulation and real-world deployment remains a central challenge in DRL for robotics. Three main strategies are commonly explored. The first relies on domain randomization, where simulation parameters are varied during training to improve robustness~\cite{Zhao2020SimtoReal}. While effective, this approach requires carefully designing parameter distributions, which can be difficult without prior knowledge of the system, and may degrade learning efficiency when overly aggressive. A second approach consists in progressively increasing simulation realism before real-world deployment~\cite{ICRA2026, Yu2019}. A third alternative consists in learning system dynamics using supervised learning from collected robot data~\cite{actuatornet, Actuatornet_dom_rand}; however, such methods often depend on the quality of the dataset, and require extensive data collection and labeling, which can limit generalization to unseen conditions. Most DRL-based robotics approaches rely either on domain randomization or on a single-simulator-to-real transfer, assuming that a sufficiently diverse or approximate simulation is enough to ensure robustness. However, this assumption often overlooks structured discrepancies between simulation engines, which we explicitly investigate here exploiting a sim-to-sim-to-real pipeline.




In light of the previous discussion, this work takes steps to extend this line of research on DRL-based maneuvering for DASMRs in two directions. First, we move from position control to full pose control, which we address by jointly adapting the state representation, reward function, and success condition. However, this extension increases task complexity and sensitivity to dynamic effects. Motivated by this fact, the second direction we investigate in this paper is analyzing the impact of ignoring actuation-related uncertainties during training, which can significantly affect policy transfer across simulators and to real systems. To this end, we adopt a sim-to-sim-to-real pipeline, where policies are trained in a simplified simulator, validated in a more realistic one, and then deployed on the physical robot. Training is performed in PyBullet due to its computational efficiency for large-scale DRL, but it relies on simplified actuation models with near-instantaneous command execution. In contrast, simulators such as Gazebo incorporate low-level controllers and capture effects such as acceleration limits, response delays, and transient behaviors. We leverage Gazebo to evaluate policies under more realistic actuation conditions without requiring full system identification. This evaluation shows that policies trained in PyBullet can suffer significant performance degradation in Gazebo. To address this issue, we propose a sim-to-sim transfer strategy where actuation effects observed in Gazebo are incorporated into the PyBullet training environment, improving the robustness of the learned policies. We demonstrate successful transfer of these policies to a real robot without additional tuning.

 \section{Problem Statement} \label{PS}
\begin{figure}[!t]
  \centering
  \captionsetup{font=scriptsize}
  \includegraphics[width=0.7\linewidth]{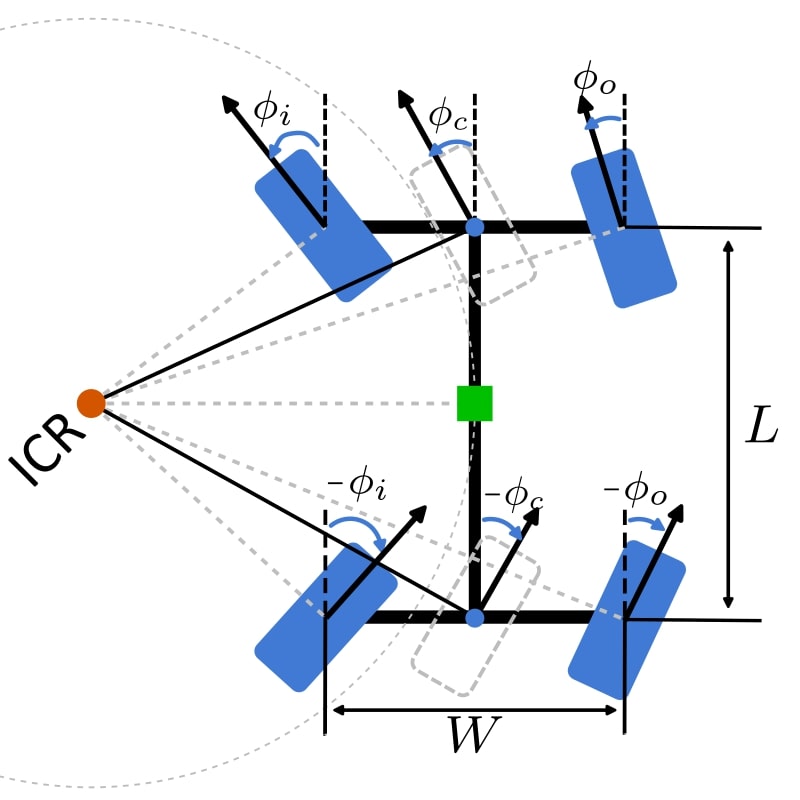}\vspace{-1pt}
  \caption{DASMR rotating around an instantaneous center of rotation (ICR).} \vspace{-0.5cm}
  \label{fig:ackermann-geom}
\end{figure}

We consider a DASMR whose current 2D pose in the world frame is given by $\mathcal{P}_c = (\boldsymbol{X}_c, \theta_c)$, where $\boldsymbol{X}_c =(x_c, y_c)$ represents the robot’s center position and $\theta_c$ the orientation of its longitudinal axis. The DASMR is tasked with reaching a desired 2D pose in the robot frame $\mathcal{P}_d = (\boldsymbol{X}_d, \theta_d)$, where $\boldsymbol{X}_d = (x_d, y_d)$ denotes the desired position for the robot's center and $\theta_d$ the desired orientation. Due to its non-holonomic nature, the DASMR cannot control all degrees of freedom independently. Its motion is restricted to forward and backward displacements coupled with orientation changes, and it cannot rotate in place. We consider a symmetric double-Ackermann configuration with a negative 4WS setup, in which front and rear steering angles are equal in magnitude and opposite in direction with respect to the chassis frame (Fig.~\ref{fig:ackermann-geom}). Although we will propose a general approach, we focus our interest on large and heavy platforms (typically above 50 kg), commonly used in agricultural applications. The robot operates within an obstacle-free workspace. The objective is to generate wheel angular velocity $\omega$ and steering angle $\phi$ commands that drive the robot toward the desired pose.

 
A key challenge is to generate feasible and robust maneuvers without relying on prior knowledge of the system dynamics or predefined motion primitives. We consider a model-free setting in which actuation dynamics, such as acceleration limits, response delays, and low-level controller effects, are not explicitly known during training. These effects introduce structured uncertainties that influence maneuver execution. While some of these uncertainties can be implicitly compensated by DRL (e.g., small tracking errors), others lead to significant performance degradation when transferring policies across simulation environments or to real-world systems. In this work, we investigate this boundary by distinguishing between tolerable and critical uncertainties, and by analyzing their impact on control performance. To address this issue, we adopt a sim-to-sim perspective. Rather than assuming a perfectly accurate simulator, we iteratively enrich the training environment by incorporating actuation effects observed in more realistic settings. This strategy allows us to reduce the gap between simplified and realistic dynamics without requiring explicit system model identification.

The control problem is formulated as a Markov Decision Process (MDP)~\cite{SuttonMIT2018}, defined by the tuple $(\mathcal{S}, \mathcal{A}, \mathcal{P}, \mathcal{R})$. At each timestep $t$, the agent observes a state $s_t \in \mathcal{S}$ and samples an action $a_t \in \mathcal{A}$ from a stochastic policy $\pi(a_t \mid s_t)$. $\mathcal{P}$ denotes the state transition function. The next state results from the interaction between the action and the environment, whose dynamics are not explicitly modeled but directly shaped through simulator-level modifications (e.g., actuation constraints such as limited acceleration or delayed response). The agent receives a reward $r_t = \mathcal{R}(s_t, a_t)$, which guides the learning process. The task is considered successfully completed when both the relative position and orientation errors in the robot frame fall below predefined thresholds.
Training is performed entirely in simulation, following common practice in DRL for robotics~\cite{Zhao2020SimtoReal}.


 \section{Method}
In our framework, a DRL agent directly controls $\omega$ and $\phi$. At the beginning of each training episode, the robot is initialized in the center of the workspace and is tasked with reaching $\mathcal{P}_d$.

\subsection{RL Background}
Actor-critic methods have shown strong performance in continuous robotic control tasks~\cite{ICRA2026, GaspardIROS2024}. These approaches rely on two neural networks: a policy (actor) and a value estimator (critic). The actor, denoted by $\pi$, defines a stochastic policy that maps a state $s_t$ to a distribution over actions, from which $a_t$ is sampled. The critic, denoted by $Q$, estimates the expected return of state-action pairs $(s_t, a_t)$ through $Q^\pi(s_t, a_t)$. The critic is trained using temporal-difference learning based on the Bellman equation~\cite{SuttonMIT2018}, with $Q^\pi_t (s_t, a_t) = r_t + \gamma \mathbb{E} [Q^\pi_{t+1} (s_{t+1}, a_{t+1})]$. The actor is updated to maximize the expected return predicted by $Q$, encouraging actions that lead to higher long-term rewards. 

\subsection{DRL Algorithms}
Our approach relies on the same DRL backbone as ManeuverNet~\cite{ICRA2026}, combining Soft Actor-Critic (SAC)~\cite{Haarnoja2018SAC} with the CrossQ framework~\cite{Bhatt2024CrossQ}. Rather than modifying the learning algorithm itself compared to ManeuverNet, we focus on improving robustness to modeling discrepancies. To this end, we employ a parallelized multi-environment training strategy, where multiple simulation instances are executed simultaneously. This setup exposes the agent to a broader diversity of trajectories and interaction patterns, which has been shown to improve generalization in robotics~\cite{CASE2022}. In our setting, it helps the policy adapt to variability induced by imperfect actuation modeling, leading to improved transfer across simulators and to real-world deployment.


\subsection{Actuation Uncertainty in Simulation} \label{sec:actuation_uncertainty}

A central aspect of this work is the study of actuation-related uncertainties and their impact on policy transfer. In DRL-based robotics, such uncertainties primarily arise from discrepancies between simplified simulation models and the real behavior of actuators. In PyBullet, actuation is typically modeled using idealized velocity control, where commanded wheel velocities are tracked almost instantaneously. This approximation neglects important physical effects such as acceleration limits and control update constraints, which can lead to unrealistic behaviors. In this work, we explicitly investigate two controllable sources of actuation uncertainty:

\textbf{(i) Acceleration limits:} In real systems, wheel velocities evolve under bounded acceleration. Ignoring this constraint leads to overly aggressive control policies that may not transfer to more realistic settings.

\textbf{(ii) Control frequency:} The rate at which control commands are updated affects the system dynamics perceived by the agent. Lower frequencies introduce discretization effects and reduce responsiveness.

These factors are directly incorporated into the PyBullet environment to progressively increase the realism of the training dynamics. In addition, we use Gazebo as a reference environment providing a more realistic representation of actuation. In this simulator, actuator behavior is governed by embedded low-level controllers, which introduce transient responses, delays, and tracking inaccuracies. While these effects are not explicitly parameterized in our study, they serve as a proxy for real-world actuation dynamics. We evaluate different configurations in Section~\ref{sec:ststr} corresponding to specific combinations of these modeling assumptions, which allows us to assess which types of actuation uncertainties critically impact policy transfer.

\subsection{State Space} 
Instead of representing the desired pose using absolute coordinates, we express the task in a relative formulation. This can facilitate generalization by encouraging the agent to learn a control strategy based on error correction rather than absolute positioning. Additionally, this formulation is consistent with the reward function in Section ~\ref{reward_new_section}, which is also defined in terms of relative errors, simplifying the learning problem. It also improves robustness to modeling discrepancies, as the policy focuses on compensating deviations rather than relying on precise absolute positioning.

More specifically, the state encodes the current robot pose $\mathcal{P}_c$ in the world frame and the desired pose $\mathcal{P}_d$ in the robot frame. In addition, the state includes low-level actuation variables. The angular spinning velocities of the left and right wheels are denoted by $\omega_l$ and $\omega_r$, while $\phi_l$ and $\phi_r$ represent the steering angles. Their corresponding steering rates are given by $\dot{\phi}_l$ and $\dot{\phi}_r$. We also include the robot body's kinematic information expressed in the world frame, namely the linear velocity of its center $\boldsymbol{V}_c = (\dot{x}_c, \dot{y}_c)$ and its angular velocity $\dot{\theta}_c$. The resulting state $\boldsymbol{s_t} \in \mathcal{S}$ observed by the agent at timestep $t$ is defined as $(\mathcal{P}_c, \mathcal{P}_d, \omega_l, \omega_r, \phi_l, \phi_r, \dot{\phi}_l, \dot{\phi}_r, \boldsymbol{V}_c, \dot{\theta}_c) \in \mathbb{R}^{15}$.

\subsection{Action Space}
The action space remains unchanged with respect to ManeuverNet~\cite{ICRA2026}.  Following the standard double-Ackermann formulation, the DRL agent outputs a  normalized control vector $\boldsymbol{a_t} = (\omega_c, \phi_c) \in [-1, 1]$, corresponding to the angular velocity and steering angle of a pair of virtual central wheels. These commands are scaled according to the robot actuation limits and then mapped  to individual wheel velocities and steering angles using the double-Ackermann geometry (Fig.~\ref{fig:ackermann-geom}). This mapping relies on the instantaneous center of rotation (ICR) and ensures consistency with the non-holonomic constraints of the platform. For brevity, the detailed derivation of the kinematic relations between $(\omega_c, \phi_c)$ and the individual wheel commands is omitted here and can be found in the ManeuverNet paper~\cite{ICRA2026}.

\subsection{Reward Function}\label{reward_new_section}
To ensure stable learning and the generation of feasible maneuvers, we build upon the reward formulation introduced in ManeuverNet~\cite{ICRA2026}. We retain the elliptical distance-based structure ($\mathcal{R}_{ES}$), originally defined using absolute coordinates in the world frame. Here, the desired pose is projected into the robot frame, yielding $(x_d, y_d, \theta_d)$, which represent the relative position and orientation of the goal with respect to the robot. The reward is defined as:
\[
r = -\sqrt{{x_d}^2 + c_1 {y_d}^2 + c_2 {\theta_d}^2},
\]
where $c_1$ and $c_2$ are weighting coefficients that balance the relative importance of lateral displacement and orientation. This choice is motivated by the DASMR kinematics, where lateral and angular deviations are typically more critical to correct than longitudinal ones. This formulation preserves the original intuition of $\mathcal{R}_{ES}$ while adapting it to a relative representation. Although the reward remains simple, the transition from position-only to pose control significantly increases task difficulty. In practice, a single maneuver is often insufficient to reach the desired pose due to dynamic constraints and modeling inaccuracies. However, the closed-loop policy enables iterative refinement, allowing the robot to perform multiple corrective maneuvers until convergence.

\subsection{Episode Termination and Success Condition} \label{episode_termination}
An episode terminates either when a success condition is met or when a maximum time horizon is reached, preventing excessively long trajectories in non-convergent cases. A trajectory is considered successful when both the relative position error $\delta_p = \|\boldsymbol{X}_d\|$ and the relative orientation error $\delta_d = |\theta_d|$ fall below predefined thresholds ${\delta_{th}}_p$ and ${\delta_{th}}_\theta$. Compared to position-only formulations, this criterion is more stringent as it requires convergence in both translation and heading.

Episodes are truncated if the robot exits the predefined workspace. This typically occurs when the robot moves too far from the desired position, preventing recovery within the episode. Such truncation ensures well-defined training episodes and avoids unbounded exploration. Due to the increased difficulty of pose control and the presence of actuation constraints, reaching the desired pose often requires multiple corrective maneuvers. This behavior naturally emerges from the closed-loop policy, which iteratively reduces the pose error until convergence.

\section{Experimental Results} \label{sec:result}

\subsection{Environment Setup}
\begin{figure}[!tb]

    \captionsetup{font=scriptsize}
    \centering 
    
    \begin{subfigure}[b]{0.75\linewidth}
        \includegraphics[width=\linewidth]{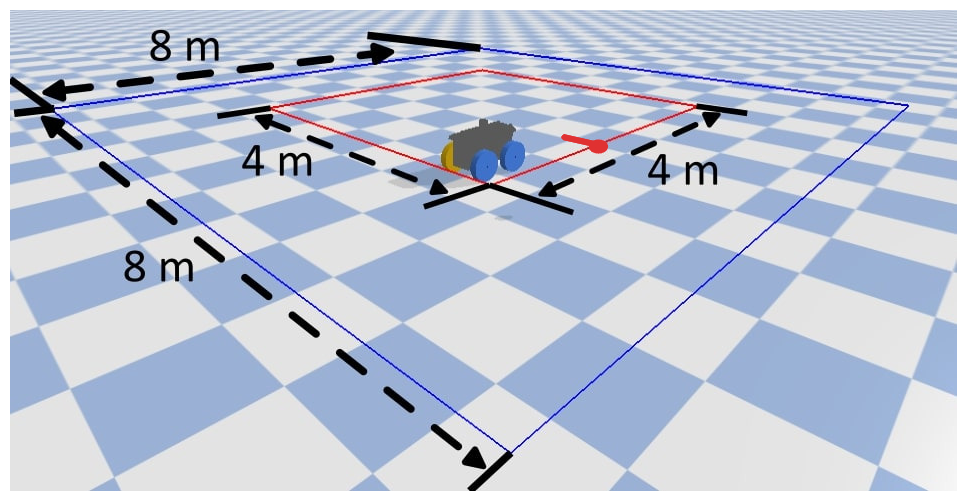}
    \end{subfigure}
    
    \vspace{0.4em}
    
    \begin{subfigure}[b]{0.75\linewidth}
        \includegraphics[width=\linewidth]{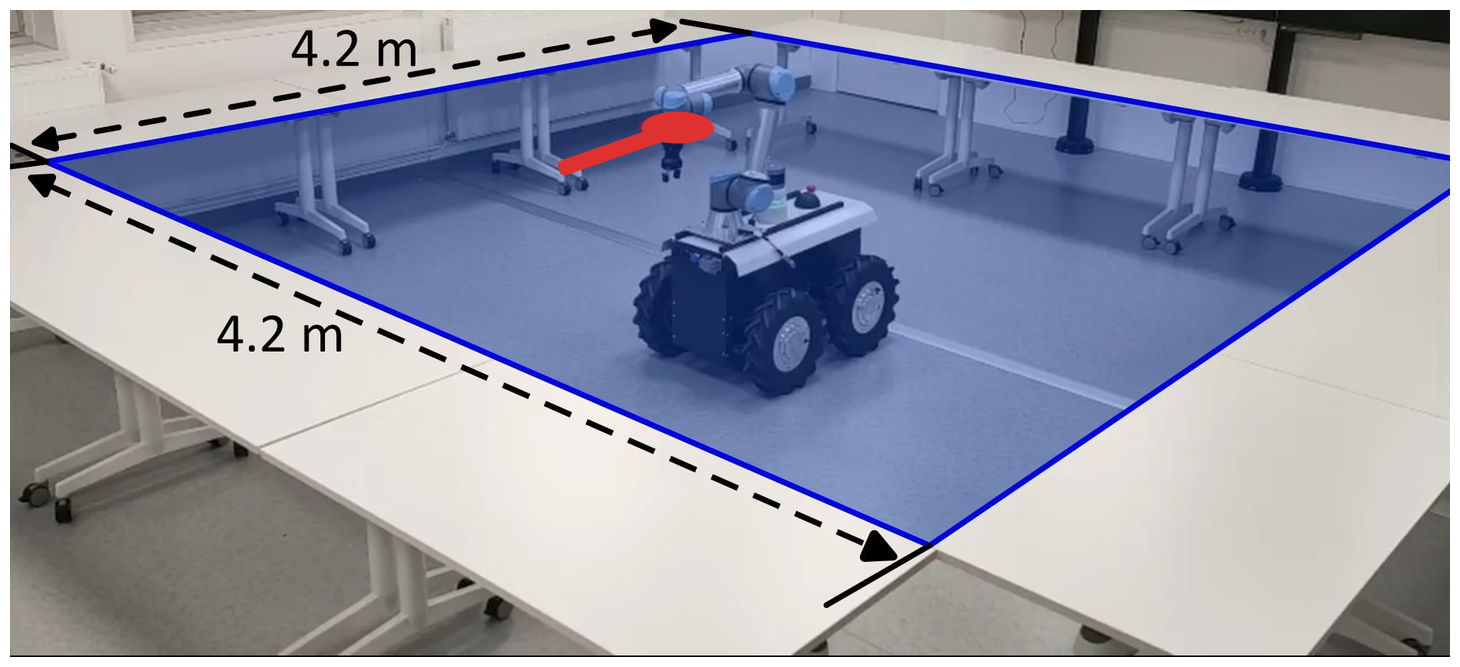}
    \end{subfigure}    
    \caption{Environment setup in simulation and real-world settings. The red square denotes the goal space. The square shaded region in blue represents the robot's workspace. In real-world settings, both spaces coincide to impose stricter constraints on navigation and positioning.} 
    \label{fig:env_spaces}%
    
\end{figure}

As in ManeuverNet~\cite{ICRA2026}, we evaluate our approach on the Shadow Runner RR100 EDU mobile robot, a 4WS platform weighing approximately 100\,kg and measuring 65\,cm in width, 90\,cm in length, and 80\,cm in height. Training is performed in the PyBullet simulation environment. The robot is initialized at the origin of an $8 \times 8$\,m$^2$ obstacle-free workspace with zero initial velocity. At each episode reset, a desired pose $\mathcal{P}_d$ is uniformly sampled within a $4 \times 4$\,m$^2$ region centered around the robot. This restricted goal space promotes the emergence of complex maneuvers requiring both accurate positioning and orientation control. For real-world experiments, the operational workspace is limited to $4.2 \times 4.2$\,m$^2$, ensuring consistency with the simulated setup while reflecting practical deployment constraints. Fig.~\ref{fig:env_spaces} illustrates both the simulation and real-world environments.

To analyze the impact of actuation modeling, we introduce an intermediate simulation stage based on Gazebo. In this environment, the robot is driven by manufacturer-provided low-level controllers, which naturally capture non-ideal effects such as acceleration limits, transient responses, and control delays. In contrast, PyBullet assumes near-instantaneous velocity tracking, resulting in simplified actuator behavior. This sim-to-sim setup enables a controlled comparison between different levels of modeling fidelity and provides a meaningful bridge toward real-world deployment.

\begin{table}[!tb]
\captionsetup{font=scriptsize}
\caption{SAC and CrossQ parameters used on SBX during training and testing.}
\scriptsize
\label{table:hyperparams}
\begin{center}
\scalebox{0.95}{
\begin{tabular}{c|c} 
\toprule
\textbf{Parameter} & \textbf{Value} \\
\hline
Nb. layers & 2 \\
Actor Hidden size & 256 \\
Critic Hidden size & 1024 \\
$\alpha_A = \alpha_C$ & 1e-3 \\
Replay buffer size & 1,000,000 \\
Batch size $N$ & 256 \\ 
$\gamma$ & 0.99 \\ 
Total training timesteps & 2,000,000 \\
Training max timesteps per episode & 800 \\
Testing max timesteps per episode & 400 \\
Training ${\delta_{th}}_p$ & 15 cm \\ 
Training and testing ${\delta_{th}}_\theta$ & $\pi/18$ \\
Training and testing random seed & 9527 \\ 
Total testing timesteps & 100,000 \\ 
Testing $d_\text{th}$ & 10, and 15 cm \\ 
\bottomrule
\end{tabular}} 
\end{center}
\end{table}

\begin{table}[!tb]
\centering
\captionsetup{font=scriptsize}
\caption{Results across PyBullet and Gazebo settings with ${\delta_{th}}_p$ in cm, the success rate (SR) in \%, and the average SPL. The three models correspond to different training regimes. ``No acceleration limits'' refers to standard PyBullet with ideal actuation. ``Model included'' includes steering acceleration limits in the simulation model. ``Lower control frequency'' reduces the control rate from 40 Hz to 20 Hz.}
\scriptsize
\label{tab:sim-to-sim}
\begin{tabular}{l | l | c | c | c}
\toprule
\textbf{Model} & \textbf{Simulator} & $\boldsymbol{{\delta_{th}}_p}$ & \textbf{SR $\uparrow$} & \textbf{SPL$\uparrow$} \\
\hline

\multirow{4}{*}{No acceleration limits} 
& \multirow{2}{*}{PyBullet} & 15 & \textbf{100} & \textbf{0.66} \\
&                           & 10 & \textbf{98} & \textbf{0.58} \\
\cline{2-5}
& \multirow{2}{*}{Gazebo} & 15 & 45 & 0.24 \\
&                         & 10 & 25 & 0.14 \\

\hline

\multirow{4}{*}{Model included} 
& \multirow{2}{*}{PyBullet} & 15 & 96 & 0.56 \\
&                           & 10 & 78 & 0.42 \\
\cline{2-5}
& \multirow{2}{*}{Gazebo} & 15 & \textbf{92} & \textbf{0.56} \\
&                         & 10 & \textbf{69} & \textbf{0.40} \\

\hline

\multirow{4}{*}{Lower control frequency} 
& \multirow{2}{*}{PyBullet} & 15 & 94 & 0.56 \\
&                        & 10 & 69 & 0.38 \\
\cline{2-5}
& \multirow{2}{*}{Gazebo} & 15 & 74 & 0.25 \\
&                         & 10 & 47 & 0.43 \\

\toprule

\end{tabular} \vspace{-0.5 cm}
\end{table}

\begin{figure*}[!tb]
    \captionsetup{font=scriptsize}
    \centering
    
    \begin{subfigure}[b]{0.19\textwidth}
        \includegraphics[width=\textwidth]{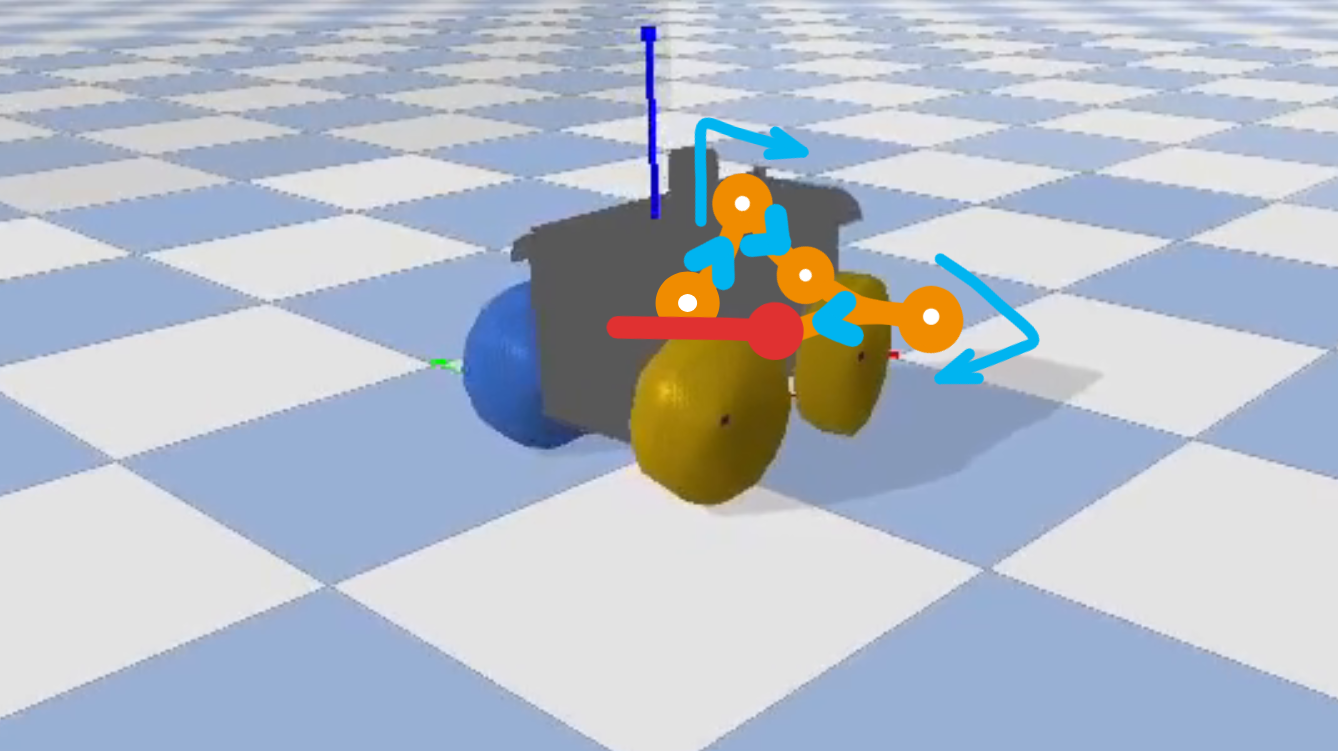}
    \end{subfigure}
    \hfill
    \begin{subfigure}[b]{0.19\textwidth}
        \includegraphics[width=\textwidth]{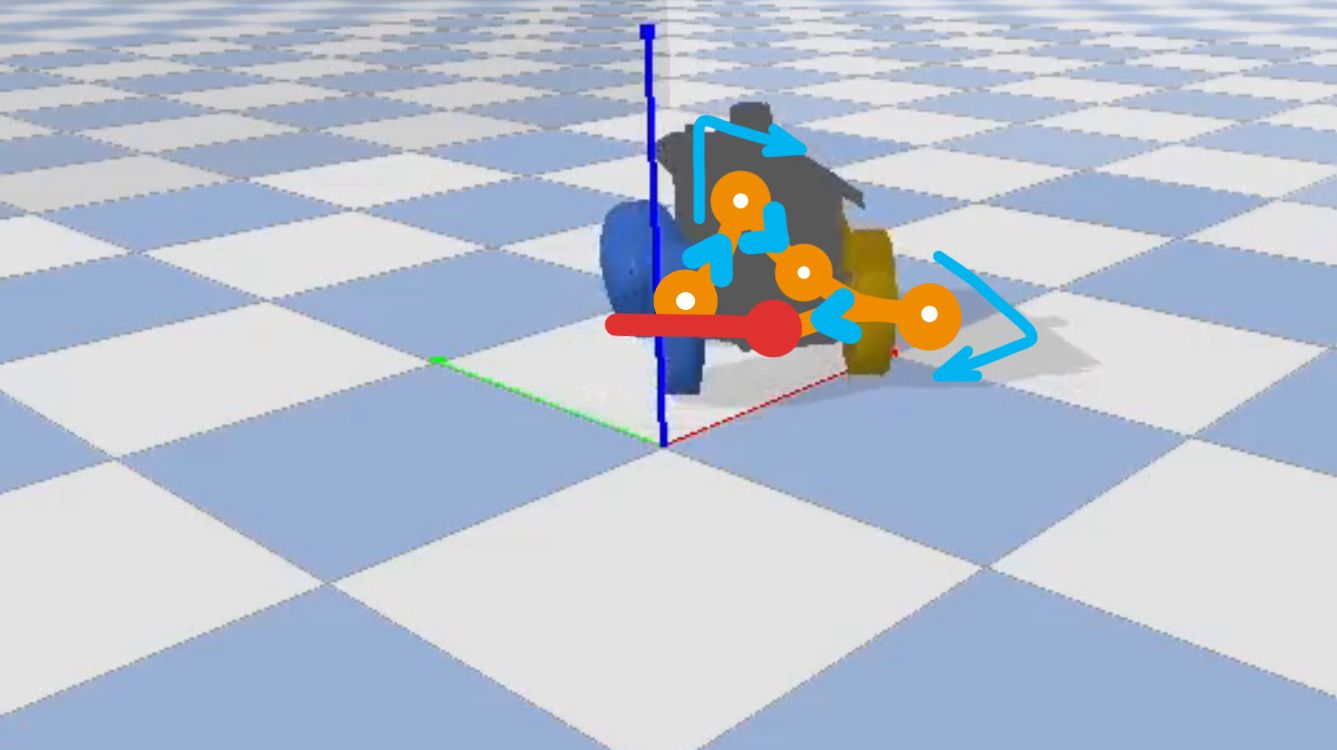}
    \end{subfigure}
    \hfill
    \begin{subfigure}[b]{0.19\textwidth}
        \includegraphics[width=\textwidth]{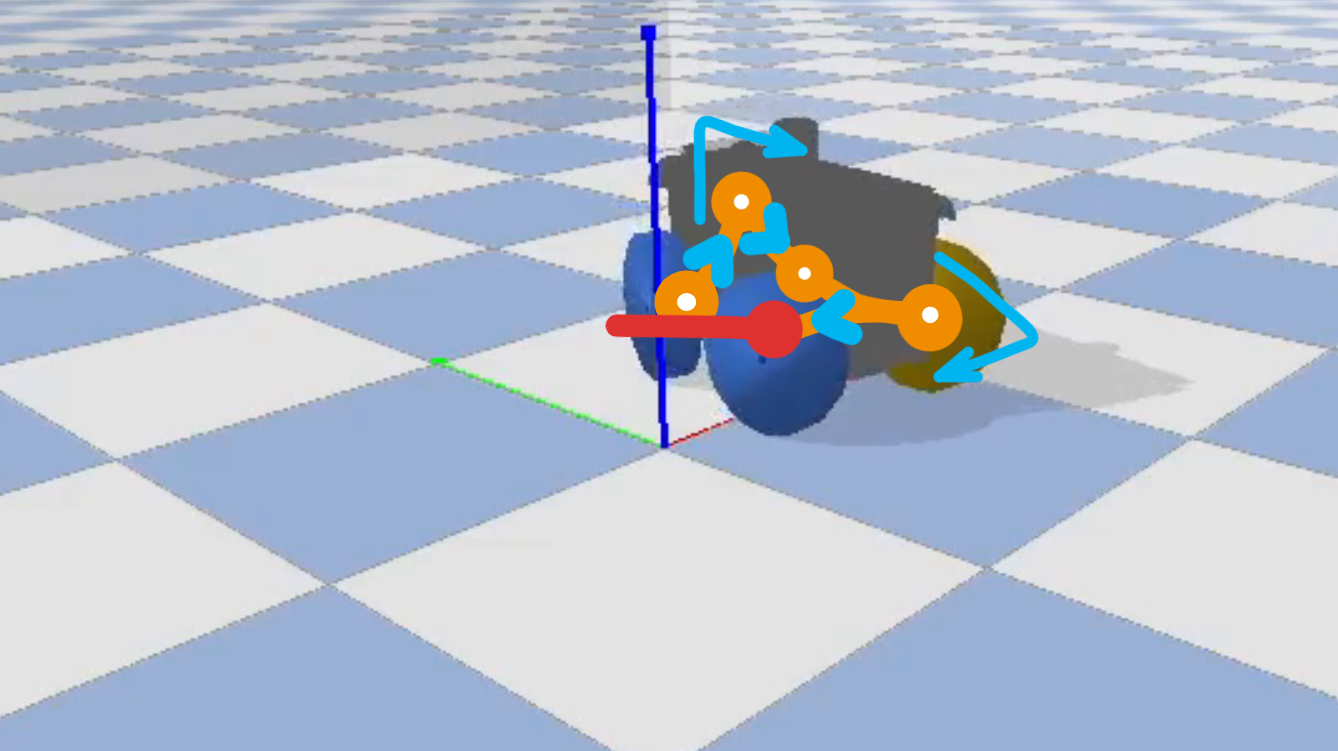}
    \end{subfigure}
    \hfill
    \begin{subfigure}[b]{0.19\textwidth}
        \includegraphics[width=\textwidth]{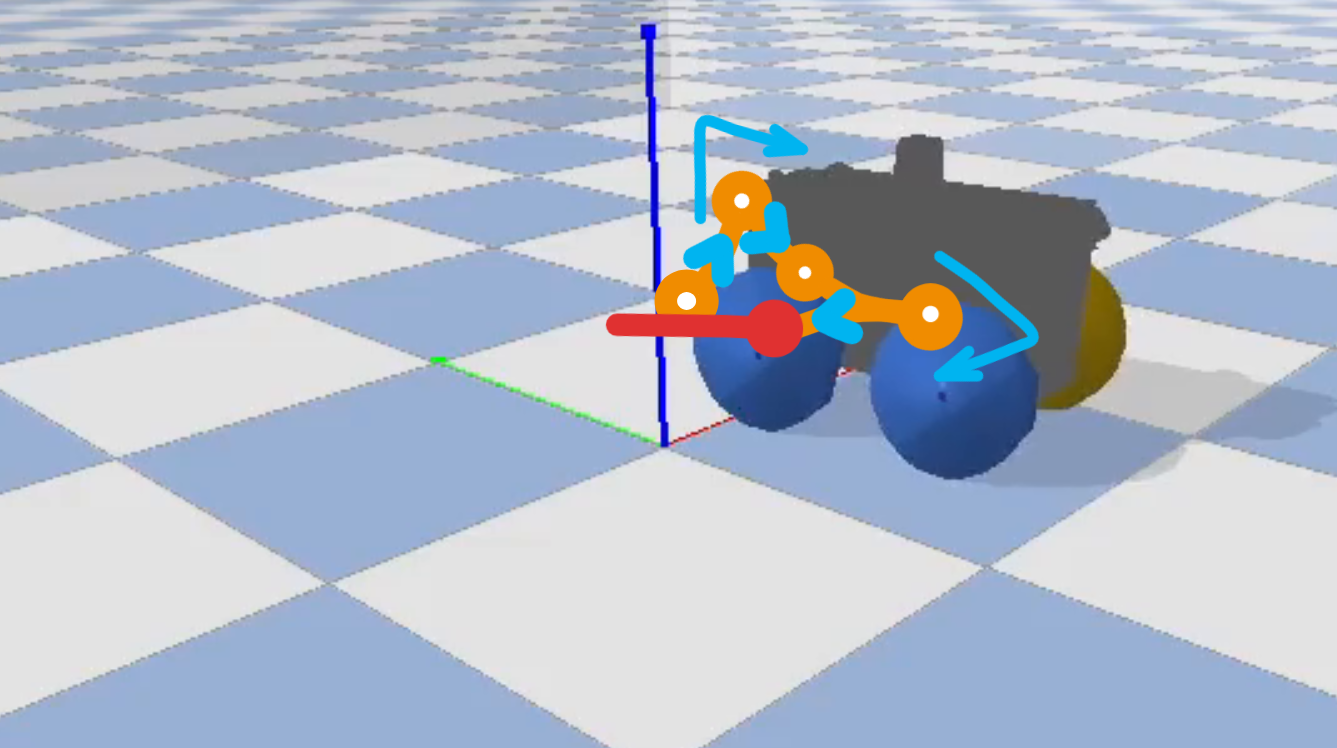}
    \end{subfigure}
    \hfill
    \begin{subfigure}[b]{0.19\textwidth}
        \includegraphics[width=\textwidth]{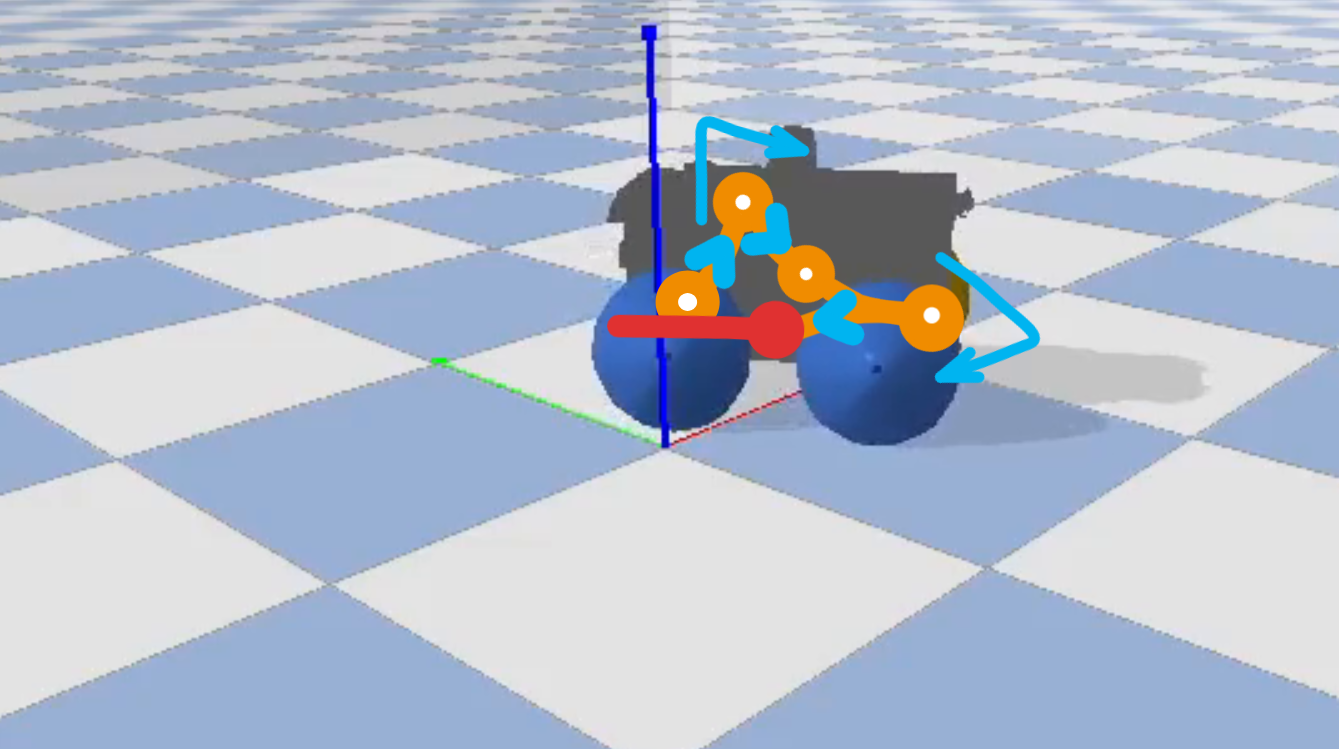}
    \end{subfigure}

    \vspace{0.2cm}

    \begin{subfigure}[b]{0.19\textwidth}
        \includegraphics[width=\textwidth]{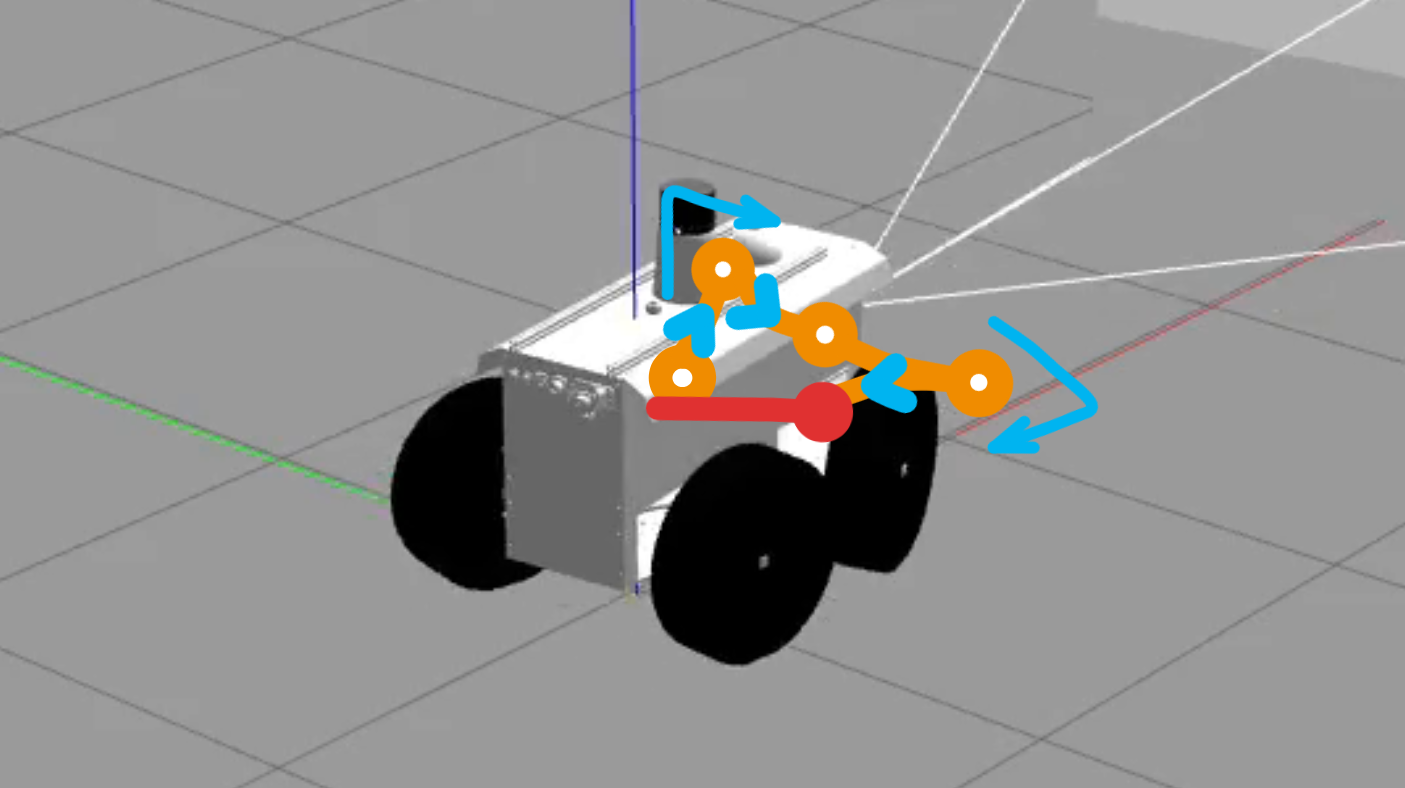}
    \end{subfigure}
    \hfill
    \begin{subfigure}[b]{0.19\textwidth}
        \includegraphics[width=\textwidth]{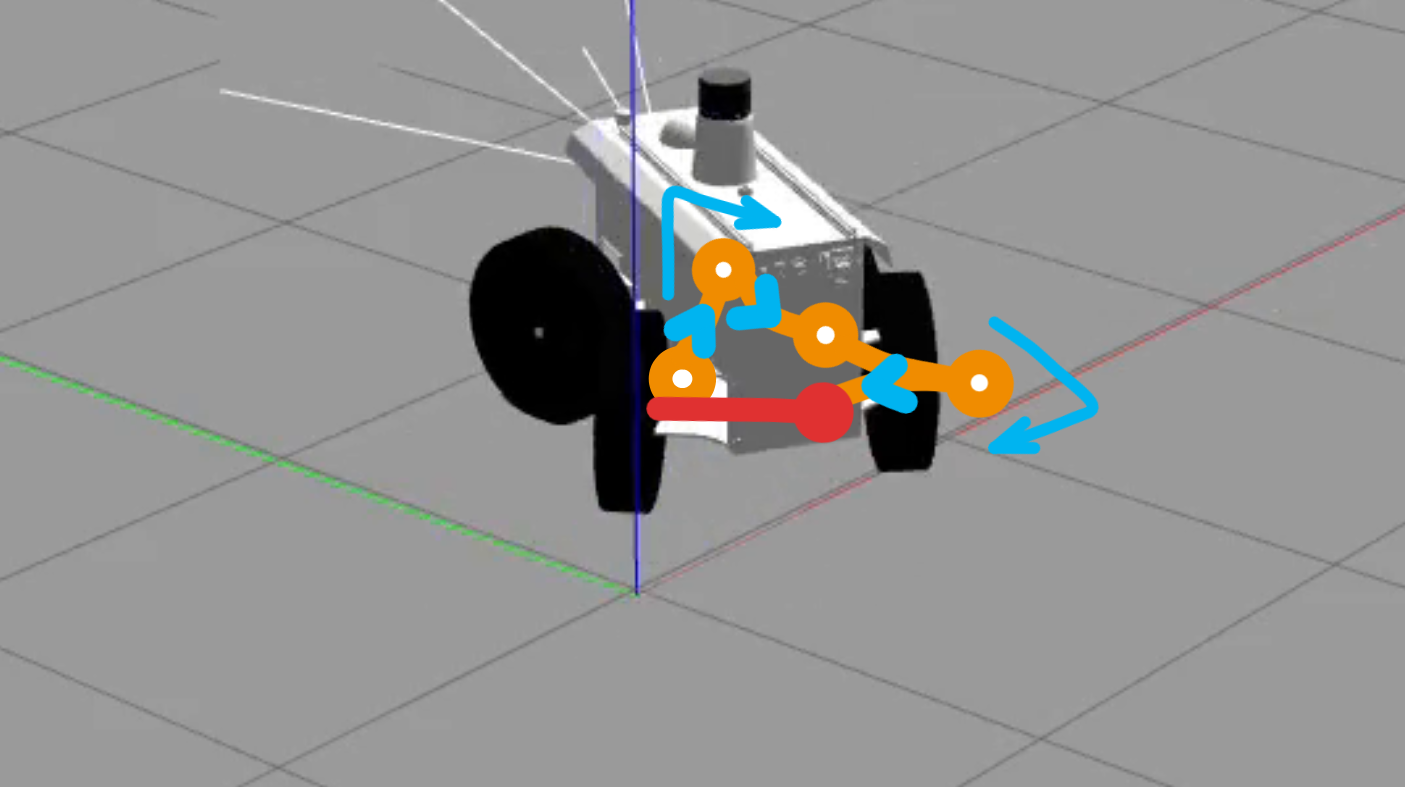}
    \end{subfigure}
    \hfill
    \begin{subfigure}[b]{0.19\textwidth}
        \includegraphics[width=\textwidth]{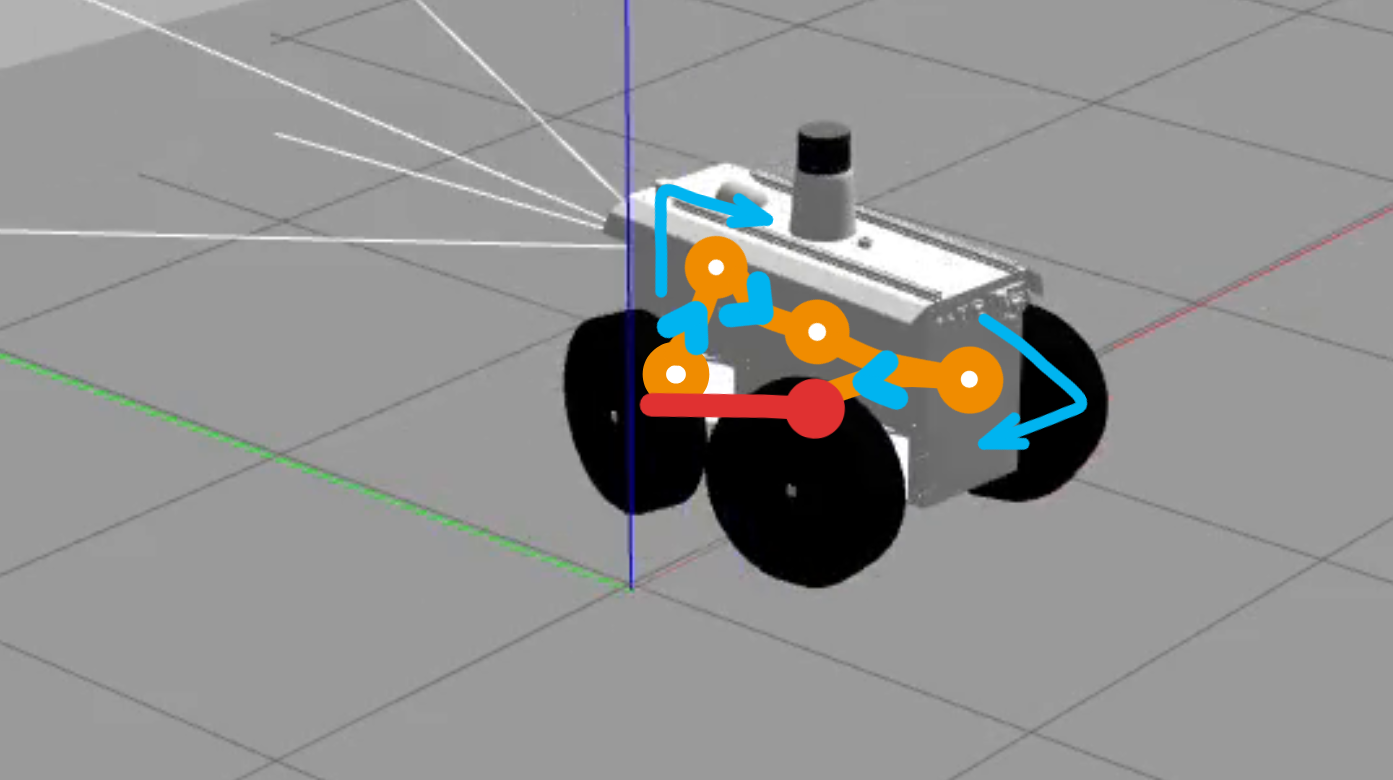}
    \end{subfigure}
    \hfill
    \begin{subfigure}[b]{0.19\textwidth}
        \includegraphics[width=\textwidth]{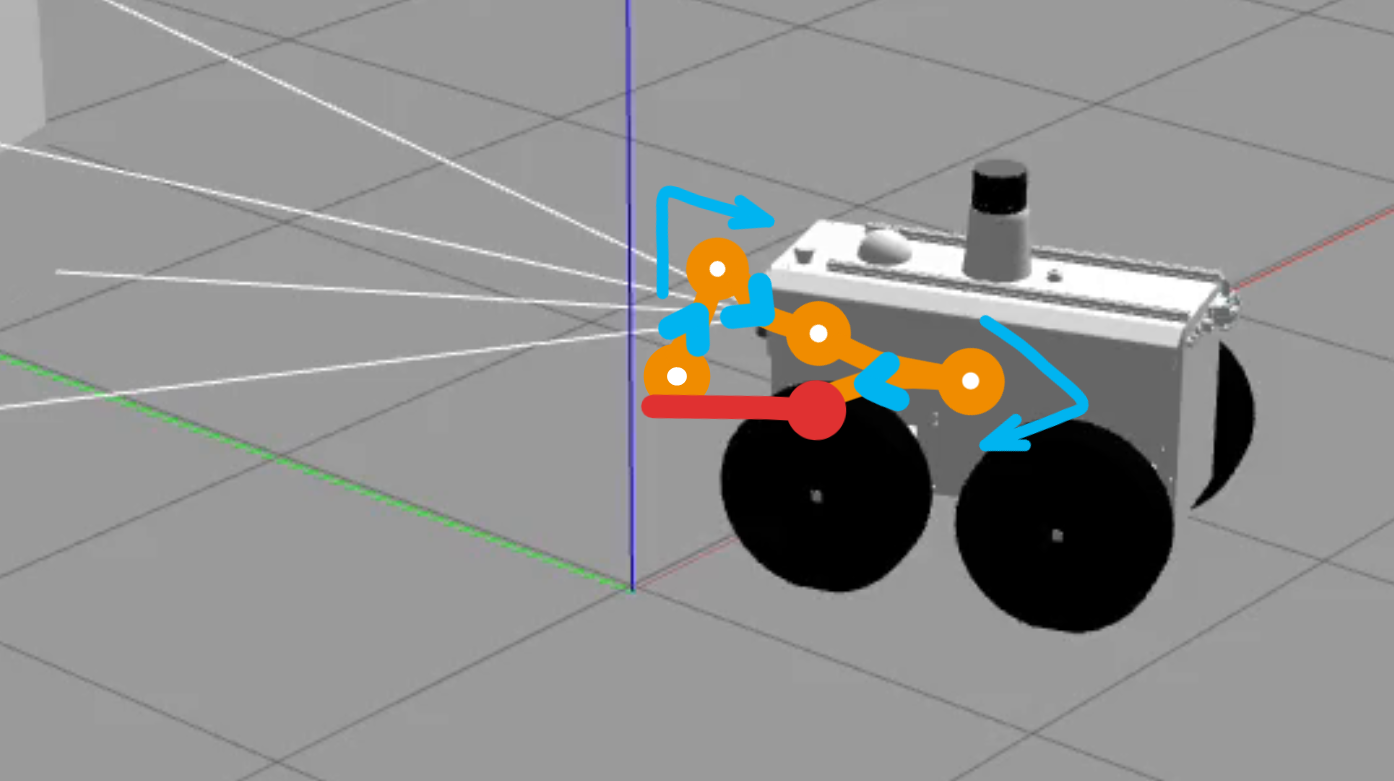}
    \end{subfigure}
    \hfill
    \begin{subfigure}[b]{0.19\textwidth}
        \includegraphics[width=\textwidth]{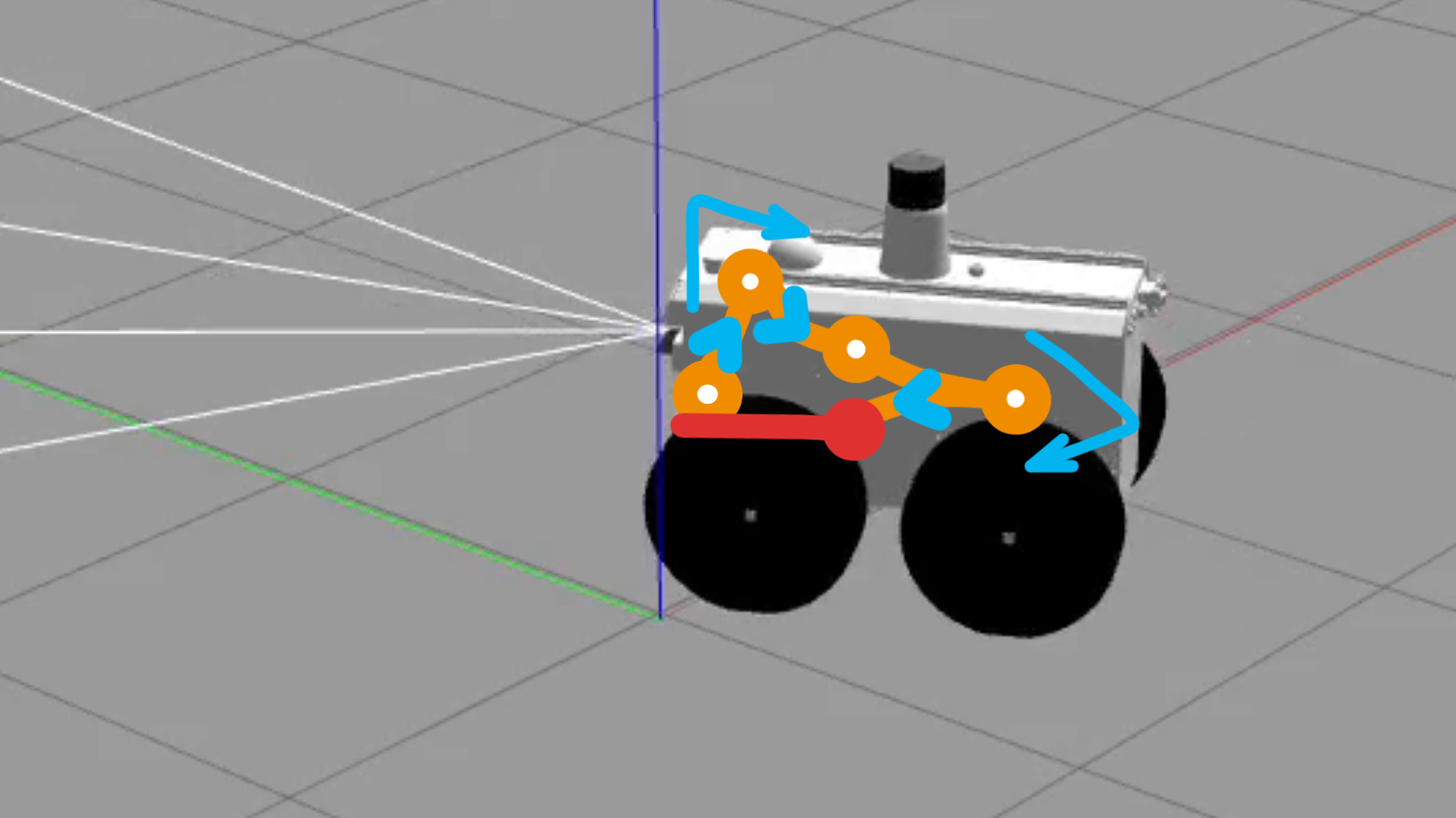}
    \end{subfigure}
    

    \vspace{0.2cm}

    \begin{subfigure}[b]{0.19\textwidth}
        \includegraphics[width=\textwidth]{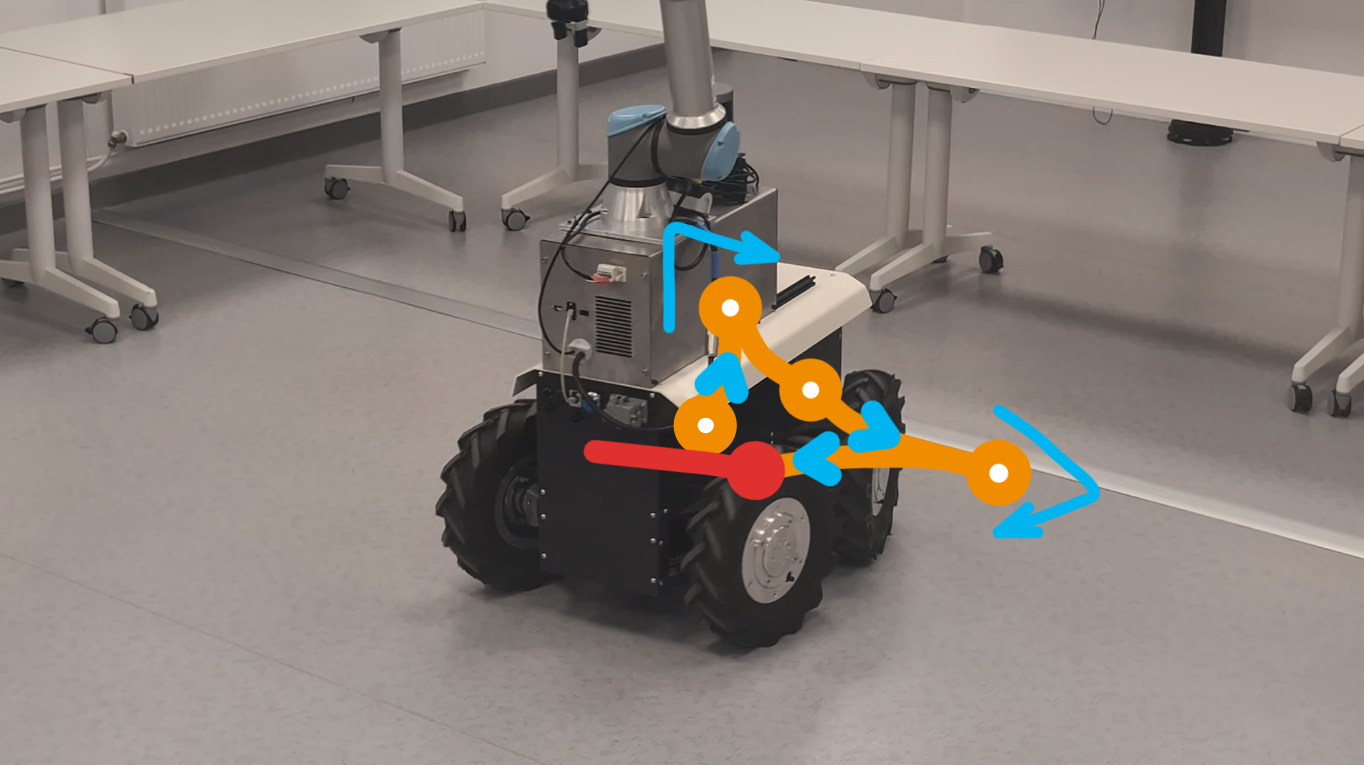}
    \end{subfigure}
    \hfill
    \begin{subfigure}[b]{0.19\textwidth}
        \includegraphics[width=\textwidth]{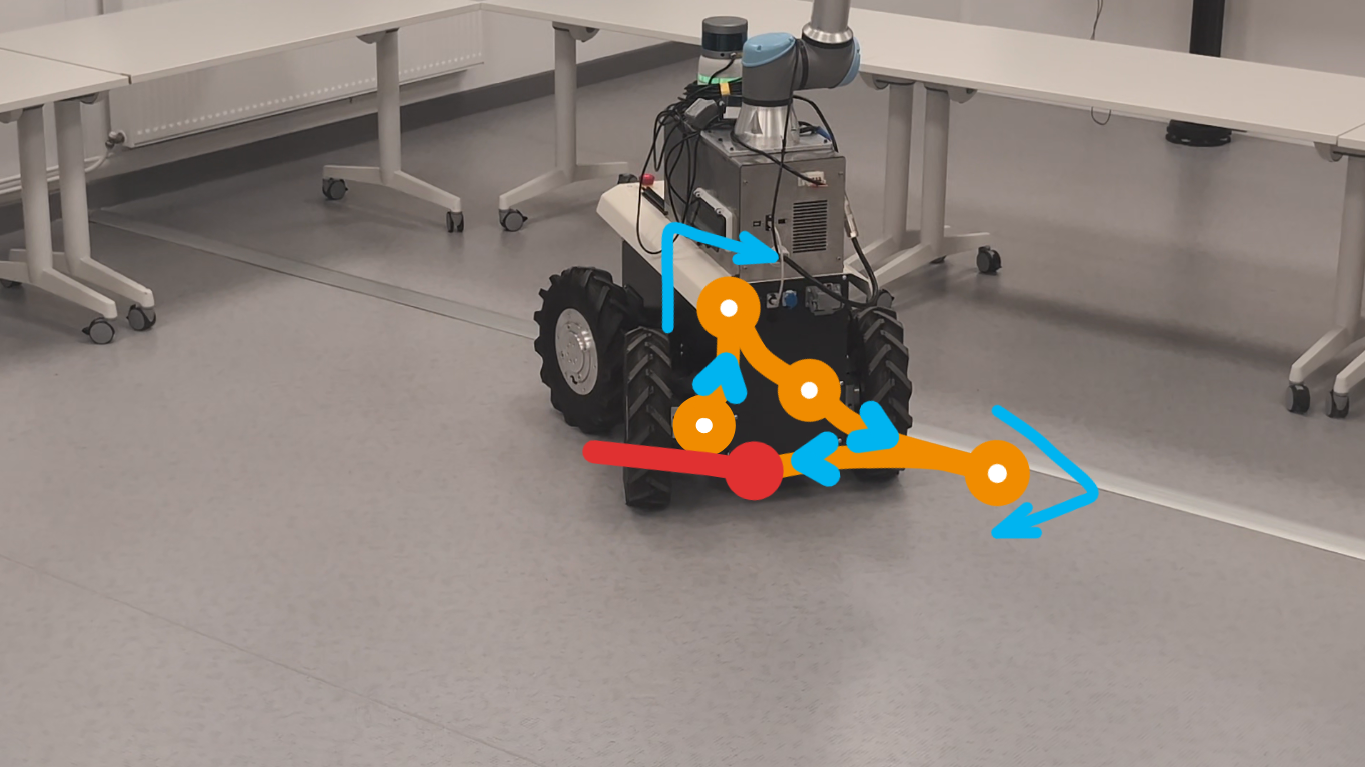}
    \end{subfigure}
    \hfill
    \begin{subfigure}[b]{0.19\textwidth}
        \includegraphics[width=\textwidth]{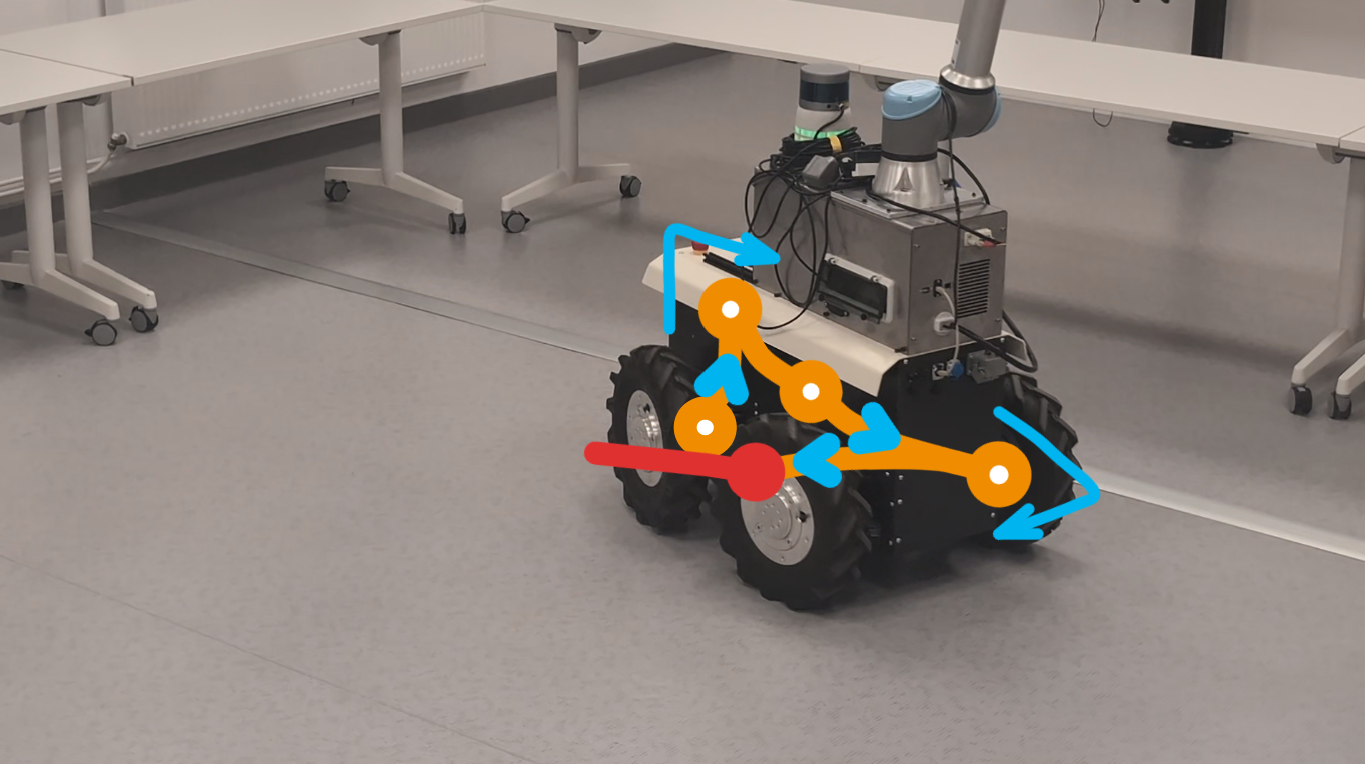}
    \end{subfigure}
    \hfill
    \begin{subfigure}[b]{0.19\textwidth}
        \includegraphics[width=\textwidth]{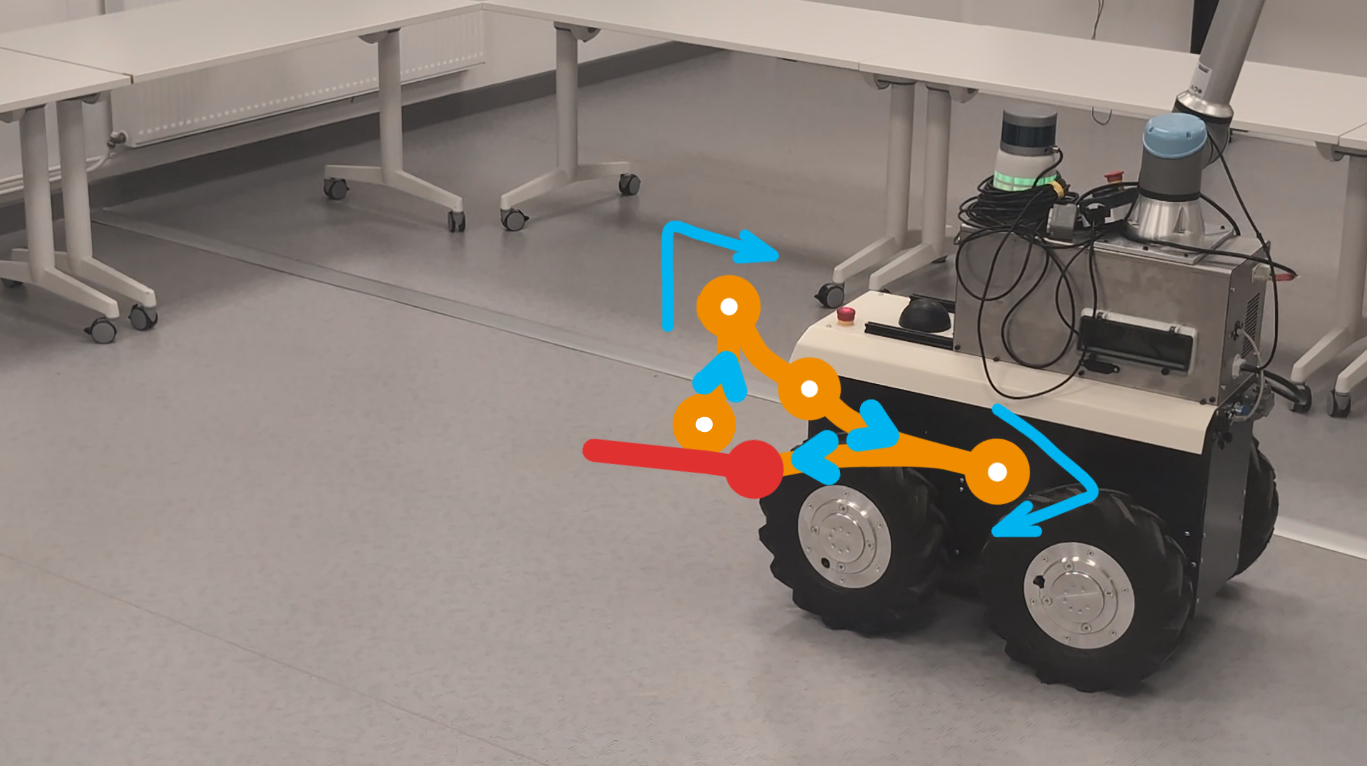}
    \end{subfigure}
    \hfill
    \begin{subfigure}[b]{0.19\textwidth}
        \includegraphics[width=\textwidth]{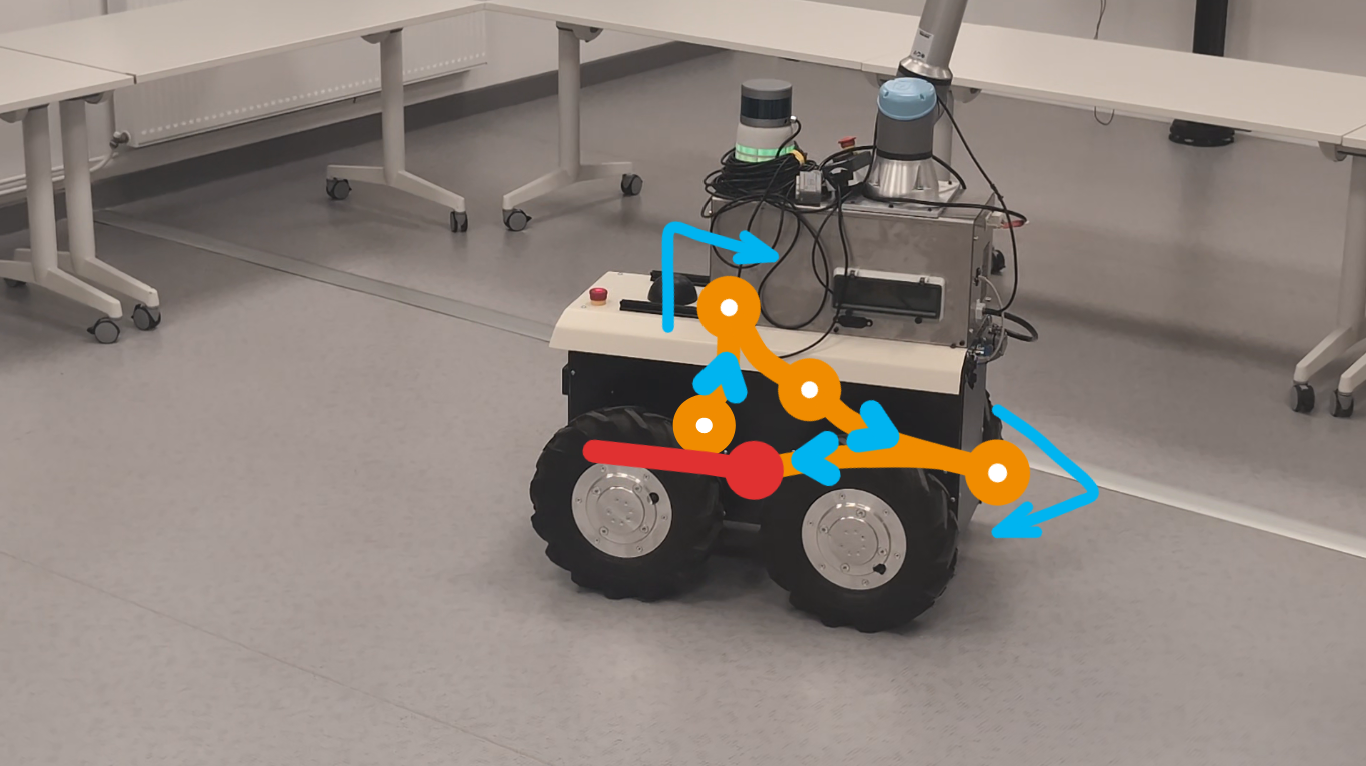}
    \end{subfigure}
    
    \caption{An example of a trajectory executed by the ``model included'' policy across PyBullet (top), Gazebo (middle), and real-world settings (bottom) is shown. The red sphere represents the desired position $\boldsymbol{X}_d$, while the red bar indicates the desired orientation $\theta_d$. From left to right in each row, the robot performs iterative corrective maneuvers under closed-loop control until convergence to the desired pose.} 
    
    \label{fig:bullet-vs-real}
\end{figure*}

\subsection{Training Setup}
The DRL agent is trained using 16 parallel simulation environments, improving data efficiency and training throughput. Training is performed over 2 million timesteps, with each episode limited to 800 steps. The control frequency is set to 40\,Hz. Under these settings, the full training process takes approximately one hour on a workstation equipped with an AMD Ryzen Threadripper PRO 7985WX (64 cores), 128\,GB of RAM, and an NVIDIA RTX 4090 GPU. All methods are implemented using the Stable Baselines JAX (SBX) library, relying on its SAC + CrossQ implementation. Unless otherwise specified, hyperparameters and network architectures are kept identical across all experiments. A detailed configuration is provided in Table~\ref{table:hyperparams}.

\subsection{Sim-to-Sim-to-Real Evaluation}\label{sec:ststr}
Our evaluation follows a three-stage pipeline designed to assess the robustness of the learned policy across simulation and real-world settings. First, the policy is trained in PyBullet, leveraging its computational efficiency for large-scale DRL. It is then evaluated in Gazebo, where more realistic actuation dynamics are introduced through low-level controllers, including acceleration limits, tracking delays, and transient response effects. Finally, policies validated in Gazebo are deployed on the real robot without additional tuning. The configurations reported in Table~\ref{tab:sim-to-sim} correspond to different levels of actuation modeling fidelity.

``No acceleration limits'' refers to the standard PyBullet configuration with ideal velocity tracking and instantaneous actuation. ``Model included'' introduces explicit wheel steering acceleration limits, resulting in non-instantaneous command tracking and bounded rate of change, while still neglecting secondary effects such as frictional losses and pressure-dependent behaviors. ``Lower control frequency'' evaluates the impact of reducing the control update rate (from 40 Hz to 20 Hz) starting from the ``Model included'' configuration, thereby introducing additional temporal discretization in action execution. These settings allow us to isolate the impact of different sources of actuation uncertainty on policy performance. An illustrative video is available at: \url{https://youtu.be/dCrjBXJ-62Q}.


We evaluate performance using the success rate (SR) and the Success weighted by (normalized inverse) Path Length (SPL)~\cite{AndersonSPL}. The success rate measures the percentage of episodes in which the robot reaches the desired pose within predefined position and orientation thresholds. SPL additionally accounts for trajectory efficiency by rewarding successful trajectories that remain close to the shortest possible path. High SPL values are particularly difficult to achieve for DASMRs, since reaching the desired pose often requires non-straight maneuvers induced by non-holonomic constraints. Quantitative results are reported in Table~\ref{tab:sim-to-sim}. Overall, they reveal a clear performance gap between simplified and high-fidelity simulation environments, and demonstrate that incorporating actuation constraints during training significantly improves transferability across simulators. We observe that when actuation dynamics are ignored, the policy achieves near-perfect performance in PyBullet (up to 100\% SR), but fails to generalize to Gazebo, where performance drops significantly (45\% SR for ${\delta_{th}}_p = 15$ cm and 25\% SR for ${\delta_{th}}_p = 10$ cm). This shows the strong sensitivity of the learned policy to actuation modeling assumptions.

In contrast, explicitly incorporating acceleration limits during training substantially reduces this gap, with strong and stable performance maintained in Gazebo (92\% SR for ${\delta_{th}}_p = 15$ cm and 69\% SR for ${\delta_{th}}_p = 10$ cm). This confirms that modeling actuator constraints plays a critical role in bridging the sim-to-sim discrepancy. Reducing the control frequency has a more moderate impact, with 74\% SR and 47\% SR in Gazebo for ${\delta_{th}}_p = 15$ cm and $10$ cm respectively. This suggests that uncertainty in temporal discretization is less detrimental than inaccurate actuation modeling. These trends remain consistent under the stricter evaluation threshold (${\delta_{th}}_p = 10$ cm), confirming that the learned policies generalize beyond the training conditions rather than overfitting to a specific tolerance. Finally, Fig.~\ref{fig:bullet-vs-real} illustrates a representative trajectory obtained with the model-aware policy. The robot performs two corrective maneuvers before reaching the desired pose, a behavior consistently observed across PyBullet, Gazebo, and real-world experiments.

 \section{Conclusion}\label{sec:conclusion}

In this work, we extended the ManeuverNet DRL framework by considering orientation in the control objective, showing that this makes it possible to learn maneuvering behaviors for pose control. We further investigated the impact of actuation-related uncertainties on policy performance. We observed that ignoring these structured discrepancies leads to a significant degradation in transfer performance, with success rate dropping from up to 100\% in PyBullet to as low as 25\% in Gazebo. In contrast, incorporating more realistic actuation dynamics during training substantially improves robustness, achieving up to 92\% success rate in Gazebo and enabling successful sim-to-sim-to-real transfer without additional tuning. As future work, we plan to address more complex scenarios, for example involving obstacle avoidance, where both perception and planning uncertainties become relevant. We also aim to conduct a more systematic study of uncertainty modeling to better understand which discrepancies require explicit modeling and which can be safely handled by learning.

\section*{Acknowledgment}
This work was supported by the French Government under the France 2030 program through the National Research Agency (ANR) grant reference ANR-24-PEAE-0002. It was also funded by the Nouvelle-Aquitaine Region through the MIRAE project. Author M. Aranda was supported through grant RYC2024-051408-I, funded by \mbox{MICIU/AEI/10.13039/501100011033} and by ESF+.

\bibliographystyle{IEEEtran}
\bibliography{biblio_iso4_abbreviations}

\end{document}